\documentclass[]{fairmeta}
\usepackage{multirow}
\usepackage{array}
\usepackage{booktabs}
\usepackage{graphicx} 
\usepackage{makecell} 
\usepackage{wrapfig}
\usepackage{subcaption} 
\usepackage{tikz}
\usepackage{pgfplots}
\pgfplotsset{compat=1.18} 
\usetikzlibrary{positioning,calc}
\usepackage[table]{xcolor}

\title{Encode, Think, Decode: Scaling test-time reasoning with recursive latent thoughts}

\author[1,2]{Yeskendir Koishekenov}
\author[2]{Aldo Lipani}
\author[1]{Nicola Cancedda}
\affiliation[1]{FAIR at Meta}
\affiliation[2]{University College London}

\abstract{
Most efforts to improve the reasoning capabilities of large language models (LLMs) involve either scaling the number of parameters and the size of training data, or scaling inference computation by letting models generate complex chains of thought. Motivated by interpretability studies showing that the crucial computation required for reasoning tasks is concentrated in a limited range of layers, we introduce Encode–Think–Decode (ETD), a method that enhances the reasoning capabilities of a base model by training it to iterate over a small subset of reasoning-relevant layers during the mid-training stage. ETD amplifies latent reasoning while preserving the original architecture, parameter count, hyperparameters, and training data composition. When iterating on the selected layers at inference time, ETD models yield substantial gains on 17 reasoning benchmarks, including +28.4\% relative accuracy improvement on GSM8K and +36\% on MATH with the OLMo-2 1B Base model. We also explore an adaptive depth strategy that adjusts the computation per input token. Our results show that recursive latent reasoning offers a simple and effective path to stronger LLM reasoning.
}

\date{\today}
\correspondence{Yeskendir Koishekenov at \email{yeskendir@meta.com}}

\begin{document}

\maketitle

\section{Introduction}
\label{section:intro}
Modern language models demonstrate remarkable capabilities in a wide range of reasoning-intensive tasks, including mathematics, programming, commonsense reasoning, and logical puzzles \citep{Brown2020LanguageMA, Dubey2024TheL3, Achiam2023GPT4TR, deepseekai2025deepseekr1}. The main driver for this progress are scale in both data and parameters, and inference-time techniques such as chain-of-thought prompting. 

Initial scaling laws correlated reasoning capabilities to sheer parameter count and training data tokens \citep{Kaplan2020ScalingLF, Hoffmann2022TrainingCL, AllenZhu2024PhysicsOL}. \citet{Ye2024PhysicsOL} refined this picture and argued that depth, not just parameter count, is critical for reasoning: deeper models often outperform shallower ones with the same number of parameters. This perspective aligns with the intuition that reasoning tasks require multi-step, compositional thinking, for which \emph{depth} plays a central role.

Beside scaling data and parameters, the prevalent approach to increasing the reasoning capability of models is by scaling test-time computation. A common approach, known as chain-of-thought (CoT) reasoning \citep{Kojima2022LargeLM, Wei2022ChainOT}, involves prompting or training LLMs to generate intermediate reasoning steps before giving a final answer.
This approach emulates human inner monologues and the use of scratchpads, but fails to capture the variability in the amount of non-verbal thought.

An emerging body of interpretability research has also sought to characterize how reasoning is implemented within LLMs. Recent studies suggest that reasoning processes are not uniformly distributed across layers, but instead transition from local, syntactic operations in earlier layers to more global and semantic integration in deeper layers \citep{Elhage2022ToyMO, Nanda2023ProgressMF, Li2022EmergentWR, Stolfo2023AMI}. Other works highlight the presence of specialized circuits and modular representations that support multi-step inference \citep{Olsson2022IncontextLA, Singh2024RethinkingII}. These findings suggest that reasoning is not merely a byproduct of scale but is tied to structured computational patterns within the network, motivating architectural modifications that amplify the contribution of reasoning-relevant layers. 

Based on these observations, we propose ETD (Encode, Think, Decode), a method to enhance the latent-space reasoning capabilities of existing models by adjusting the effective depth of the network. We identify a range of critical layers for latent reasoning and train it into becoming a recurrent block.

Recursive depth models, also known as looped models, have been mostly studied as a way to improve parameter efficiency \citep{Lan2019ALBERTAL, Bae2024RelaxedRT}. Our goal in applying a recursive approach, conversely, is to boost reasoning capabilities by efficiently scaling inference-time computation. There has been work on measuring the effectiveness of recursive-depth models on fairly simple reasoning tasks \citep{saunshi2025reasoning}, and deliberate attempts to improve reasoning via such looping \citep{geiping2025scaling}. However, these works apply recursion without explicitly targeting the layers most relevant for reasoning within the model.

Rather than training small models from scratch to compare recursive and non-recursive variants, we validate our approach on pretrained open-source models from the OLMo 2 family \citep{OLMo20242O2}. We re-run their mid-training stage to integrate recursion, but crucially, we do not introduce additional parameters, new data, or changes to the original hyperparameters. This makes our method practical and straightforward to reproduce, as it builds on widely available pretrained models without requiring costly retraining from scratch. To our knowledge, this is the first work to demonstrate that introducing recurrent depth yields significant improvements over modern open-source LLMs.

We demonstrate that our proposed method leads to significant improvements across 17 tasks requiring different types of reasoning. Notably we achieve a relative improvement of 28.4 \% and 36\% on GSM8K \citep{Cobbe2021TrainingVT} and MATH \citep{Hendrycks2021MeasuringMP} for the OLMo-2 1B base model. 

We also propose how to dynamically set the depth of the model depending on the token. This allows to spend less compute on easy problems and more compute on challenging ones.

The main contributions of the paper are as follows:
\begin{itemize}
    \item We show that advanced open-source pretrained models can be further enhanced with a recurrent-depth mechanism that requires no additional parameters, training data, or hyperparameter tuning.
    \item We demonstrate that ETD provides greater benefits on tasks requiring intensive reasoning, with relative improvements of 28.4\% on GSM8K and 36\% on MATH for OLMo-2 1B.
    \item We analyze the impact of iterating over different layers on reasoning performance and introduce a practical recipe for selecting critical layers for latent reasoning.
    \item We show that performing more latent-space reasoning, i.e. increasing the number of iterations, directly improves performance on reasoning tasks.
    \item We introduce a mechanism to adaptively determine the number of iterations for each input. 
\end{itemize}

\section{On the Roles of Layers for Reasoning}
\label{sec:layer_roles}
There have been extensive studies on the functional roles of different layers in neural networks. In computer vision, shallow layers are known to capture general features, while deeper layers represent more fine-grained ones \citep{Zeiler2013VisualizingAU, Bau2017NetworkDQ}. Similar patterns are also observed in LLMs. For example, \citet{Stolfo2023AMI} show that, when solving simple arithmetic questions, LLMs encode information about operators and operands in mid-sequence early layers, transform this information into intermediate computations in middle layers, and form the representation of the final answer in the last-token middle-to-late layers. Likewise, \citet{Zhao2024LayerBL} find that, during instruction tuning, early layers capture broad and reusable knowledge, middle layers amplify task-relevant signals, and deeper layers refine these signals into task-specific outputs. More broadly, interpretability studies confirm functional differentiation across layers of varying depths, including in reasoning settings \citep{Yu2025BackAU, gromov2024unreasonable, shi2024understanding, Skean2025LayerBL}. 

As information propagates from early to deeper layers, the reasoning process transitions from specific, local, and syntactic information  to rich semantic integration. We draw the conclusion that early to middle layers play a critical role in task understanding \citep{davidson2025different} and knowledge retrieval, while deeper layers are important for higher-level inferences such as those required for mathematical reasoning.

\begin{figure}[t]
\centering
\begin{subfigure}[t]{0.5\textwidth}
    \centering

    \includegraphics[width=\linewidth]{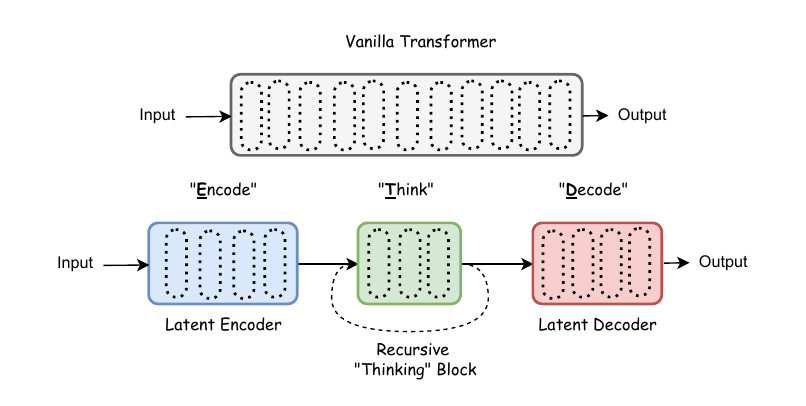}
    \label{fig:etd_diagram}
\end{subfigure}
% \hfill
\begin{subfigure}[t]{0.4\textwidth}
    \centering
    \raisebox{1.25em}{%
    \includegraphics[width=\linewidth]{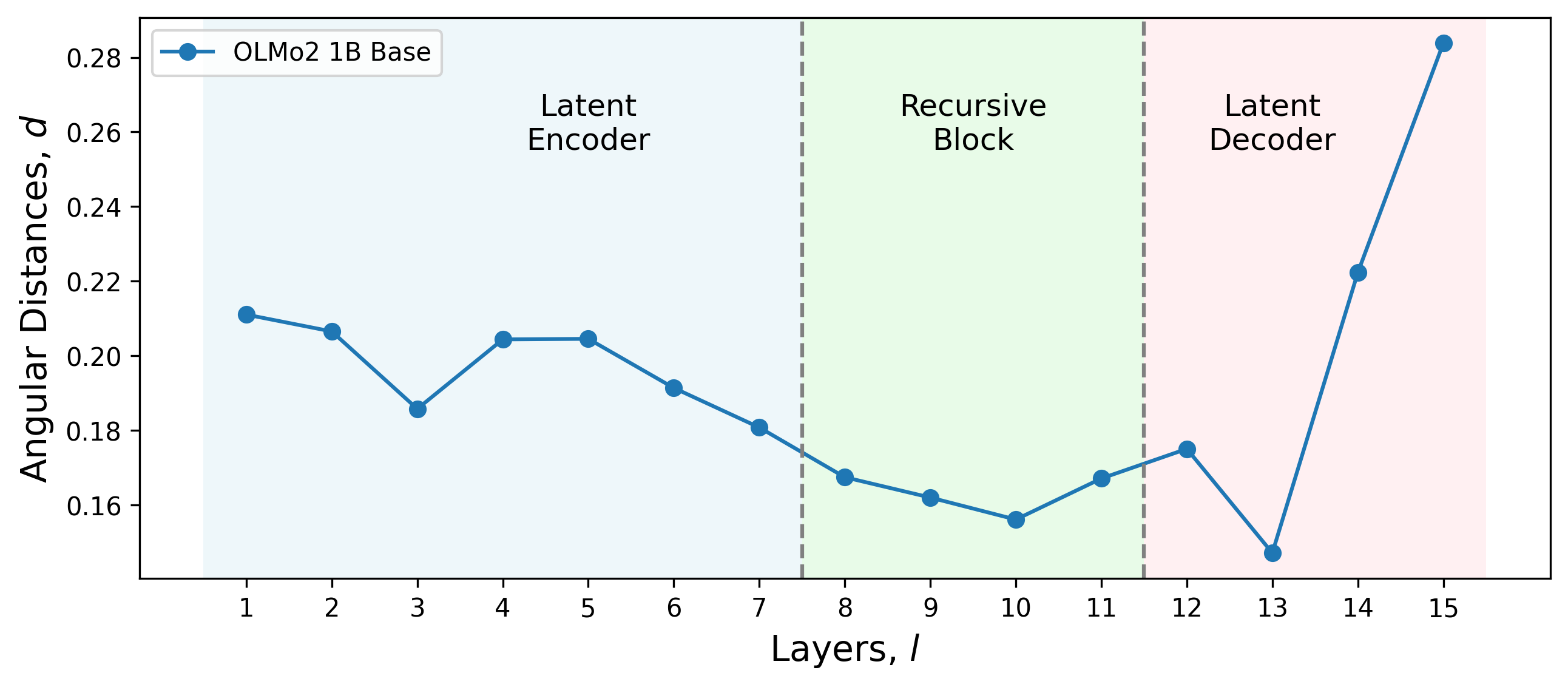}
    }
    \label{fig:layer_sim}
\end{subfigure}
\vspace{-2.0em} % adjust amount as needed
\caption{\textit{Left}: Illustration of the proposed architecture (Section~\ref{sec:choosing_config}). The latent encoder (blue) maps inputs into latent space, the recursive “thinking” block (green) iteratively refines representations, and the latent decoder (red) maps them back to the output space. Each block consists of a different number of layers.
\textit{Right}: Angular distances $d(l, l+1)$ between consecutive layers for OLMo 2 1B base model. The plot highlights three groups of layers—latent encoder, recursive block, and latent decoder—corresponding to distinct trends in layer-to-layer evolution (Section~\ref{sec:choosing_config}).}
\label{fig:combined}
\end{figure}

We therefore break down transformer blocks into three groups (Figure \ref{fig:combined}): a latent encoder $E$, which embeds the input data into a latent space and retrieves information about mentioned entities, then a core recurrent ``thinking'' block $T$, a central unit of recurrent computation, that generates latent ``thoughts'', and finally the latent decoder $D$, which un-embeds from latent space and also contains the prediction head of the model. In practice, the information first goes through layers in the latent encoder $E$, then iterates over the ``thinking'' block $k$ times, and finally flows through the latent decoder $D$, which returns output tokens. 
Let's denote different configurations as $ N_{E}$-$ N_{T}$*$k$-$ N_{D}$, e.g. 7-4*2-5 denotes a transformer with 7 layers in the $E$ block, 4 layers in the $T$ block, repeated twice, and 5 layers in the $D$ block.

If the layer-to-layer evolution of representations is given by a residual iteration equation:
\begin{equation}
    x^{l+1} = x^{l} + f(x^{l}, \theta^{l})
\end{equation}
where $x^{l}, \theta^{l}$ are the input and parameter vectors for layer $l$, and $f(x^{l}, \theta^{l})$ represents the transformation of one multi-head self-attention and MLP layer block \citep{Vaswani2017AttentionIA}, then after $L$ total layers the output is the sum of the input embeddings and the contributions of all the layers:
\begin{equation}
    x^{L} = x^{0} +
    \sum_{l=0}^{N_{E}-1}f(x^{l}, \theta^{l}) +
    \sum_{j=1}^{k}\sum_{l=N_{E}}^{N_{E}+N_{T}-1}f(x^{l+(j-1)*N_{T}}, \theta^{l}) + 
    \sum_{l=N_{E}+n_{T}}^{L-1}f(x^{l+(k-1)*N_{T}}, \theta^{l})
\end{equation}

\subsection{Choosing the optimal configuration for latent reasoning}
\label{sec:choosing_config}

Prior work on related recursive architectures largely relied on \textit{ad hoc} design choices. Some approaches apply recursion over all internal layers, i.e. employ only a recursive block $T$, \citep{Dehghani2018UniversalT, Csordas2024MoEUTMU, Bae2024RelaxedRT, saunshi2025reasoning}, others allocate 1–2 layers each to the $E$ and $D$ blocks \citep{geiping2025scaling, Bae2025MixtureofRecursionsLD, Aleksandrov2025AbbIEAB}. In contrast, our work takes the roles of layers into consideration when determining the configuration.

The latent encoder should include enough layers to transform input text into the latent space and retrieve all relevant knowledge, laying the foundation for higher-level semantic analysis and reasoning to happen via a recursive ``thinking'' block, $T$.

To identify the optimal configuration of layers, we build on the approach of \citet{gromov2024unreasonable}.
They discovered that later layers change the direction of hidden representations less than earlier layers. They used the average angular distance as a criterion for identifying layers to prune. Their experiments show that removing such layers has almost no impact on tasks heavily relying on knowledge retrieval. Despite the low average angular change, however, even moderate pruning of those same layers results in a degradation on reasoning tasks. We build on these insights and use mean angular change to identify reasoning-critical layers to iterate over.

We measure the average change in the direction of the residual stream vector after each layer, and add layers to the latent encoder until the rate of change from layer to layer slows down.

In practice, we compute the average angular distance $d(x(l), x(l+n))$\footnote{We explain the details of computing angular distance in Appendix \ref{app:angular_distance}}, between the input to layer $l$ and the input to layer $l+n$ on the C4 validation set \citep{Raffel2019ExploringTL}. The distance quantifies the degree of update to $x$ resulting from processing between layers $l$ and $l+n$. Figure \ref{fig:combined}(right) shows the average distances $d(x(l), x(l+1))$ for OLMo-2 1B base and instruct models.

To automatically identify the point, i.e. the layer, at which a curve transitions from a rapid to a gradual decrease, we employ the \textit{Kneedle algorithm} \citep{Satopaa2011FindingA}. This method detects ``knee'' (or ``elbow'') points in convex, decreasing sequences by analyzing their curvature. Algorithm details are provided in Appendix \ref{sec: kneedle_algorithm}. The detected layer index defines the boundary of the latent encoder. For the OLMo-2 1B model, this corresponds to layer 7.

Similarly to the latent encoder, the latent decoder must have sufficient depth to transform representations from the latent space back into the “language” space. To determine the number of layers in the latent decoder, we follow the same procedure as for the latent encoder, but applied in reverse: starting from the final layer of the model and moving backward until reaching the last layer assigned to the latent encoder. For the OLMo-2 1B model, this yields the last 5 layers as the latent decoder. The remaining 4 layers constitute the recursive ``thinking'' block.

Hence, we set the configuration to 7-4*k-5, i,e. 7 layers in latent encoder, 4 layer in recursive  block, and 5 layers in latent decoder respectively, and k is number of iterations. In Figure \ref{fig:combined} (right), the rate of change in angular distance decreases around layer 7, stabilizes over the subsequent four layers, and increases again during the final five layers.

\section{Experimental Setup}

Prior works on recursive-depth models typically rely on simplified training setups. We are, however, interested in understanding the impact of recursive ``thinking'' in realistic scenarios, and therefore apply them on open-source models trained following best practices in architecture, training recipe, and pretraining data mixtures. We base our study on the OLMo 2 family of models \citep{OLMo20242O2}, focusing specifically on the base configurations. For fair comparison, our ETD models use the same number of parameters, datasets, and hyperparameters as the baseline non-recursive model.

\subsection{Training Pipeline}
\label{sec:training_pipeline}

OLMo 2 is a family of LLMs with open artifacts including intermediate and final checkpoints, training data, code, and recipes for 1B, 7B and 13B scale models, both pre-trained and post-trained. As a compromise between experimental agility and model power, we focus on 1B parameter model. We integrate ETD into the existing training pipeline without introducing additional training steps or data. This requires access to the model weights, training data, and hyperparameters to evaluate the impact of ETD in a controlled and isolated manner.

Following recent advances in curriculum learning \citep{Blakeney2024DoesYD, Ibrahim2024SimpleAS} OLMo 2 base models are trained in two stages. The first (pretraining) stage is the longest ($\geq 90\%$ training FLOPs), and uses mostly web-sourced data. The second stage, which is referred to as mid-training (5-10 \% of training FLOPs), upsamples the highest-quality web documents and curated non-web sources. The purpose of this mixture is to imbue the model with reasoning skills and provide focused exposure to STEM references and high quality text.

We evaluate the ETD approach by integrating it into the mid-training stage which uses only 1.25\% of the total pretraining tokens.\footnote{For the OLMo-2 1B model, stage-1 pretraining uses $4\times 10^{12}$ tokens, while stage-2 uses $5\times 10^{10}$ tokens.} In our experiments, we initialize the model with the weights after the first stage training and run the mid-training with ETD approach for each configuration separately. \citet{OLMo20242O2} perform mid-training with three random orders, then average the resulting models. In our setup, we train with one data configuration and compare it to the standard model trained with the same configuration. Since our experiments adopt the same data mixtures and configurations, we direct readers to \citet{OLMo20242O2} for a comprehensive description of the training pipeline.

\subsection{Evaluation Benchmarks}
\label{sec:evaluation_metrics}

\begin{wraptable}{r}{0.55\textwidth} 
\vspace{-12pt}
\centering
\caption{Evaluation benchmarks grouped into six categories, listed in order of increasing reasoning intensity from top to bottom.}
\label{tab:evaluation_benchmarks}
\resizebox{0.55\textwidth}{!}{
\begin{tabular}{ll}
\toprule
\textbf{Category} & \textbf{Benchmarks} \\
\midrule
Factual Knowledge & TriviaQA, NaturalQuestions \\
\cmidrule(lr){1-2} % \addlinespace
Reading Comprehension & BoolQ, OpenBookQA, DROP \\
\cmidrule(lr){1-2} % \addlinespace
\multirow[t]{2}{*}{Commonsense Reasoning} 
& CommonSenseQA, HellaSwag \\
& SocialQA, WinoGrande \\
\cmidrule(lr){1-2} % \addlinespace
\multirow[t]{2}{*}{Multi-Disciplinary Reasoning} 
& ARC-Easy, ARC-Challenge, MMLU, \\
& MMLU-Pro, AGIEval-English \\
\cmidrule(lr){1-2}
BIG-Bench Hard & BBH \footnotemark \\
\cmidrule(lr){1-2} % \addlinespace
Mathematical Reasoning & GSM8K, MATH \\
\bottomrule
\end{tabular}
}
\end{wraptable}
\footnotetext{BBH, a collection of 23 diverse tasks, serves as a cross-cutting benchmark for compositional reasoning that does not fit neatly into the other categories. More details in Appendix \ref{app:eval_benchmarks}}

To capture broad conceptual nature of reasoning, we consider 17 real-world benchmarks grouped into six categories, ordered along a spectrum from less to more reasoning intensive tasks, i.e. from factual recall to systematic symbolic reasoning: factual knowledge, reading comprehension, commonsense reasoning,  multi-disciplinary Reasoning, BIG-Bench Hard (BBH), and mathematical reasoning. This progression reflects increasing reliance on reasoning rather than memorization. We provide the task categories with the corresponding benchmarks in Table \ref{tab:evaluation_benchmarks}. Details with the motivation for each task category are provided in Appendix \ref{app:eval_benchmarks}. We evaluate the model using OLMES \citep{Gu2024OLMESAS}, a standardized evaluation suite and toolkit.

\begin{table}[ht]
\caption{Results of the Encode–Think–Decode (ETD) method with varying numbers of iterations over recursive “thinking” blocks, compared to the OLMo 2 1B baseline. Reported metrics include accuracy (Acc.) and relative improvement ($\Delta$, in \%) with respect to the baseline, for each of six task categories (as defined in Sec. \ref{sec:evaluation_metrics}). Parameter counts denote the number of distinct layers, while FLOPs correspond to the number of effective forward-pass layers.}

\label{tab:looping_results}
\centering
\renewcommand{\arraystretch}{1.2}
\setlength{\tabcolsep}{5pt} 

\resizebox{\textwidth}{!}{
\begin{tabular}{r c| *{5}{c c|}c c}

\toprule
\noalign{\vskip 0.5ex}  
& 
& 

\multicolumn{2}{c}{\makecell{Factual\\Knowledge}} & 
\multicolumn{2}{c}{\makecell{Reading\\Comprehension}} & 
\multicolumn{2}{c}{\makecell{Commonsense\\Reasoning}} & 
\multicolumn{2}{c}{\makecell{Multi-Disciplinary\\ Reasoning}} & 
\multicolumn{2}{c}{BBH} & 
\multicolumn{2}{c}{\makecell{Math.\\Reasoning}} \\
% [0.75ex] 
\cmidrule(lr){3-4} \cmidrule(lr){5-6} \cmidrule(lr){7-8} \cmidrule(lr){9-10} \cmidrule(lr){11-12} \cmidrule(lr){13-14}
Model & Params/FLOPs & \multicolumn{1}{c}{Acc.} & $\Delta (\%)$ & Acc. & $\Delta (\%)$ & Acc. & $\Delta (\%)$ & Acc. & $\Delta (\%)$ & Acc. & $\Delta (\%)$ & Acc. & $\Delta (\%)$ \\
\midrule
OLMo 2 (k=1) & 16 / 16  & 37.55 & - & 52.19 & - & 65.29 & - & 45 & - & 31.8 & - & 24.31 & - \\
\rowcolor{gray!10}
ETD (k=2) & 16 / 20 & 38.1 & (+1.5\%) & 56.14 & (+7.6\%) & 66.74 & (+2.2\%) & 48.41 & (+7.6\%) & 31.67 & (-0.4\%) & 28.27 & (+16.3\%) \\
\rowcolor{gray!5}
ETD (k=3) & 16 / 24 & 37.55 & (0\%) & 56.07 & (+7.4\%) & 67.75 & (+3.77\%) & 49.55 & (+10.1\%) & 32.62 & (+2.6\%) & 30.29 & (+24.6\%) \\
\rowcolor{gray!10}
ETD (k=4) & 16 / 28  & 37.74 & (0\%) & 57.76 & (+10.7\%) & 68.16 & (+4.4\%) & 50.18 & (+11.5\%) & 33.01 & (+3.8\%) & 29.62 & (+21.8\%) \\
\rowcolor{gray!5}
ETD (k=5) & 16 / 32 & \textbf{38.23} & (\textbf{+1.8\%}) & \textbf{58.5} & (\textbf{+12.1\%}) & \textbf{68.41} & (\textbf{+4.8\%}) & \textbf{50.58} & (\textbf{+12.4\%}) & \textbf{33.49} & (\textbf{+5.3\%}) & \textbf{30.45} & (\textbf{+25.3\%}) \\
\bottomrule
\end{tabular}
}
\end{table}

\section{Evaluating Recursive ``Thinking'' Blocks}
All results are obtained using the training pipeline described in Section \ref{sec:training_pipeline}, with the only modification being the configuration $N_{E}$-$N_{T}$*$k$-$N_{D}$. Here, $N_{E}$, $N_{D}$, and $N_{T}$ denote the number of layers in the latent encoder and decoder, and the recursive block, and $k$ is the number of iterations. Since our objective is to evaluate the model’s reasoning abilities, we focus on reasoning-oriented tasks as defined in Section \ref{sec:evaluation_metrics}. Because we deal with the same architecture while changing only the number of layers, we report the number of parameters in terms of distinct layers, $N_{E}$+$N_{T}$+$N_{D}$, and the number of FLOPs in terms of forward passes through layers, $N_{E}$+$N_{T}$*$k$+$N_{D}$. 

\subsection{Performance Gains from Iterating over ``Thinking'' Blocks}

We begin by examining the first two rows of Table \ref{tab:looping_results}, which report results for the baseline and the recursive model with two iterations, corresponding to the 7–4*2–5 configuration. Notice that the  OLMo 2 1B parameter baseline is equivalent to the  ETD model with $k$=1. Results show that performance either remains stable or improves, with notable gains in several categories. The largest improvement is observed on Mathematical Reasoning tasks, with an average relative increase of 16.3\%. A breakdown in Table \ref{tab:math_results} confirms that both GSM8K and MATH benefit from two iterations of the ETD approach. Additional gains appear in Commonsense Reasoning (+2.2\%), Reading Comprehension (+7.6\%), and Multi-Disciplinary Reasoning (+7.6\%). In contrast, tasks in the Factual Knowledge and BIG-Bench Hard categories exhibit at most marginal benefits from a single additional iteration. These findings motivate further exploration of the ETD approach with more iterations, which we examine next. Detailed results for all 17 tasks are provided in Appendix \ref{app:etd_all_tasks}.

\subsection{Scaling Behavior of Latent Reasoning}

\begin{wraptable}{r}{0.5\textwidth}
    \vspace{-8pt}
    % \centering
    
    \caption{Results of the ETD method with varying numbers of iterations. Reported metrics include accuracy (Acc.) and relative improvement ($\Delta$, in \%) with respect to the baseline on the mathematical reasoning tasks, GSM8K and MATH.} 
    \setlength{\tabcolsep}{5pt}
    \resizebox{0.5\textwidth}{!}{
    \begin{tabular}{r c|*{1}{c c|}c c}
        \toprule
        \noalign{\vskip 0.5ex}
        & & \multicolumn{2}{c}{GSM8K} & \multicolumn{2}{c}{MATH} \\
        \cmidrule(lr){3-4} \cmidrule(lr){5-6}
        Model & Params/FLOPs & Acc. & $\Delta (\%)$ & Acc. & $\Delta (\%)$  \\
        \midrule
        OLMo 2 (k=1) & 16 / 16  & 44.05 & - & 4.57 & - \\
        \rowcolor{gray!10}
        ETD (k=2) & 16 / 20 & 51.10 & (+16.01\%) & 5.45 & (+19.22\%)  \\
        \rowcolor{gray!5}
        ETD (k=3) & 16 / 24 & 54.36 & (+23.41\%) & \textbf{6.22} & (\textbf{+36.04\%})  \\
        \rowcolor{gray!10}
        ETD (k=4) & 16 / 28 & 55.50 & (+25.99\%) & 3.73 & (-18.28\%) \\
        \rowcolor{gray!5}
        ETD (k=5) & 16 / 32 & \textbf{56.56} & (\textbf{+28.4\%}) & 4.33 & (-5.17\%)   \\
        \bottomrule
    \end{tabular}
    }
    \label{tab:math_results}
    \vspace{-15pt}
\end{wraptable}

To further assess the effect of recursive processing, we train ETD with varying numbers of iterations, the results summarized in Table \ref{tab:looping_results}. Performance generally improves as the number of iterations $k$ increases.

The main exception is the Factual Knowledge category with negligible improvement. As discussed in Section \ref{sec:evaluation_metrics}, these tasks rely mainly on memorization rather than reasoning. In contrast, the largest gains occur in reasoning-intensive tasks, most notably in Mathematical Reasoning, with breakdowns shown in Table \ref{tab:math_results}.

These results demonstrate that the ETD approach—by iterating over reasoning-relevant layers—substantially enhances the non-recursive baseline, yielding relative improvements of +28.4\% on GSM8K and +36\% on MATH. Moreover, the minimal gains on memorization tasks further validate our approach from Section \ref{sec:layer_roles} for identifying layers specialized in reasoning.

As noted earlier, ETD with $k$=2 iterations shows no improvement on BIG-Bench Hard (BBH) tasks. However, performance begins to increase with $k$=3 and continues to improve with additional iterations. These observations highlight that performance as a function of iterations exhibits different trends across tasks. For some tasks (e.g., Social IQa, ARC-Challenge, MMLU), performance rises rapidly with 2–3 iterations, after which the rate of improvement slows. For others (e.g., DROP, MMLU-Pro, GSM8K), gains continue steadily with each additional iteration. In rare cases, the best performance is not achieved at the maximum depth, as observed for MATH. 

Overall, these findings indicate that allocating more resources to generating latent ``thought'' before decoding—that is, by performing additional iterations over the ``thinking'' blocks—systematically enhances performance on reasoning-oriented tasks. The diverse performance trends across tasks highlight the opportunity to explore input-dependent, adaptive-depth recursive methods, which we investigate in Section \ref{sec:adaptive_depth}.

\subsection{Comparison with alternative recursive frameworks}
Prior work on recursive LLMs typically applies recursion either across all layers \citep{Dehghani2018UniversalT, Csordas2024MoEUTMU, Bae2024RelaxedRT, saunshi2025reasoning} or across middle layers while preserving a few initial and final layers \citep{geiping2025scaling, Bae2025MixtureofRecursionsLD, Aleksandrov2025AbbIEAB}. For a fair comparison, we train models using both strategies: (i) looping over all layers, and (ii) a 2–12*2–2 configuration, which repeats the middle 12 layers while keeping two layers at the beginning and end fixed. We compare these baselines to our selective looping configuration under a constant FLOP budget, with results shown in Table~\ref{tab:looping_baselines}. 

Our approach consistently outperforms these alternatives under equal compute. For example, the 2–12*2–2 setup is FLOP-equivalent to our 7–4*4–5 configuration, yet yields lower accuracy. Moreover, to match or exceed the performance of alternative strategies, our method typically requires fewer FLOPs—often only three iterations are sufficient.

\begin{table}[h]
\caption{Results with recursive baselines}

\label{tab:looping_baselines}
\centering
\renewcommand{\arraystretch}{1.2}
\setlength{\tabcolsep}{4pt} 

\resizebox{0.7\textwidth}{!}{%
\begin{tabular}{c|c|*{5}{c |}c}
\toprule
\noalign{\vskip 0.5ex}  
\multirow{1}{*}{Model} & 
\multirow{1}{*}{\makecell{Params/\\FLOPs}} & 
% \multirow{2}{*}{FLOPs} & 
\multicolumn{1}{c|}{\makecell{Factual\\Knowledge}} & 
\multicolumn{1}{c|}{\makecell{Reading\\Comprehension}} & 
\multicolumn{1}{c|}{\makecell{Commonsense\\Reasoning}} & 
\multicolumn{1}{c|}{\makecell{Multi-Disciplinary\\ Reasoning}} & 
\multicolumn{1}{c|}{BBH} & 
\multicolumn{1}{c}{\makecell{Math.\\Reasoning}} \\
[1.75ex] 

\hline
OLMo 2 & 16 / 16  & 37.55  & 52.19  & 65.29  & 45  & 31.8 & 24.31  \\
\midrule
2-12*2-2 & 16 / 28 & 37.7  & 56.44  & 67.73  & 47.58  & 32.30  & 29.27  \\
\rowcolor{gray!10}
ETD (k=4) & 16 / 28 & \textbf{37.74}  & \textbf{57.76}  & \textbf{68.16} & \textbf{50.18}  & \textbf{33.01}  & \textbf{29.62}  \\
\midrule
0-16*2-0 & 16 / 32  & 37.35  & 53.58  & 64.7  & 45.24  & 30.59  & 24.99  \\
\rowcolor{gray!10}
ETD (k=5) & 16 / 32 & \textbf{38.23}  & \textbf{58.5}  & \textbf{68.41}  & \textbf{50.58}  & \textbf{33.49} & \textbf{30.45} \\
\bottomrule
\end{tabular}
}
\end{table}

\subsection{How does the choice of recursive layers change performance?}

\begin{figure}[ht!]
\begin{center}
%\framebox[4.0in]{$\;$}
% \fbox{\rule[-.5cm]{0cm}{4cm} \rule[-.5cm]{4cm}{0cm}}
\includegraphics[width=\textwidth]{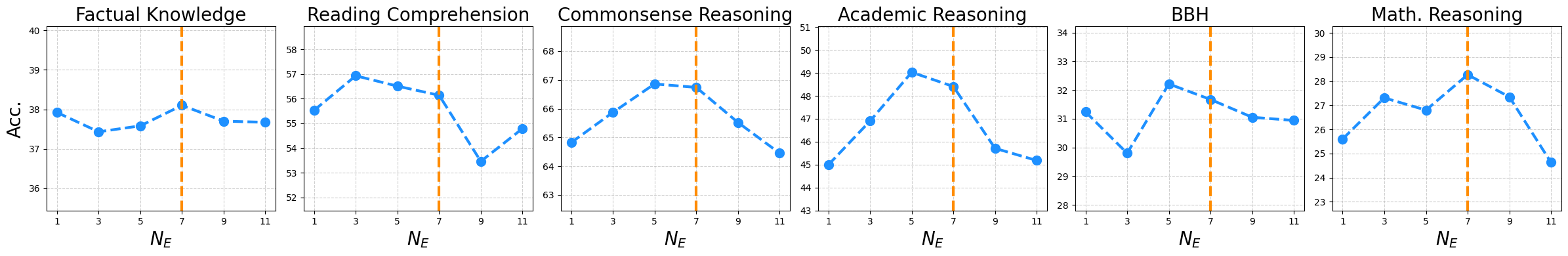}
\end{center}
\caption{Results of the ETD method when varying the subset of layers in the recursive block. We report accuracy (Acc.) when increasing the size of the latent encoder $N_{E}$ from 1 to 11 in steps of 2, for each of 6 task categories (as defined in Sec. \ref{sec:evaluation_metrics}). The orange line marks selected configuration.}
\label{fig:moving_T_plot}

\end{figure}

To further examine the impact of recursive layer choice, we fix the recursive ``thinking'' block size and vary its starting position from layer 2 to 12 in steps of 2, which is equivalent to increasing the size of the latent encoder $N_{E}$ from 1 to 11 in steps of 2. An intriguing observation is that the optimal configuration slightly varies depending on the specific category of tasks. The results in Figure \ref{fig:moving_T_plot}  show that the 7-4*2-5 configuration achieves the best overall performance on reasoning-intensive task, particularly mathematical reasoning\footnote{Exact values can be found in Table \ref{tab:looping_layers} in Appendix \ref{app:looping_layers}}. A close alternative is 5-4*2-7, which performs comparably on most tasks but falls short in mathematics. Performance on Factual Knowledge tasks is stable across configurations, which aligns with the intuition discussed earlier. Interestingly, for reading comprehension, the 3-4*2-9 configuration performs best. This block of layers (4-7) overlaps with layers just before the identified “thinking” block (8-11), aligning with our earlier intuition that early-to-middle layers are important for context understanding. These findings are consistent with our layer-role analysis, though further investigation is needed to establish stronger causal links. To conclude, our results empirically demonstrate that the framework described in Section \ref{sec:layer_roles} enables the selection of configurations that enhance the model’s reasoning capabilities.

\section{Adaptive test-time scaling}
\label{sec:adaptive_depth}
We observed significant improvements of iterating over recursive blocks. The general trend is that the model benefits from more iterations. However, different problems demand different levels of reasoning effort: not all tokens or sequences require the same number of iterations to reach an accurate prediction, and in some cases the marginal benefit of additional iterations may not justify the extra computation. Adaptive computation \citep{Bengio2013EstimatingOP, Bengio2015ConditionalCI} is often used for efficiency by early-exiting on simpler tokens \citep{Elhoushi2024LayerSkipEE}. In contrast, our goal is to adaptively allocate computation at test time to enhance reasoning capability, rather than to reduce cost.

\subsection{Methodology}

In our architecture of the form $E\!\to\! T*k\!\to\! D$, instead of fixing the number of recursive iterations $k$, we adopt the Adaptive Computation Time (ACT) mechanism \citep{Graves2016AdaptiveCT}, allowing each token to dynamically determine how many applications of the recursive block $T$ are necessary. A lightweight \emph{router} evaluates the hidden state after each iteration and decides whether further computation is required. This enables allocating more steps to tokens that demand deeper reasoning, while those not meeting the selection criteria bypass further processing and retain their previous representation.

At each iteration $t$, after computing the hidden representation $h_t$ with the recursive block, a \emph{router} predicts a halting values $w_t \in (0,1)$ for each token. These values are accumulated across iterations:
\begin{equation}
    H_t = \sum_{j=1}^t w_j.
\end{equation}
Computation for a token is stopped once $H_t \geq 1 - \epsilon$, with $\epsilon$ is a small constant (e.g. 0.01), and the maximum possible number of iterations is $N_{max}$=10. Intuitively, each $w_t$ represents the confidence of the latent ``thought'', as produced by the recursive block $T$. Until sufficient confidence is accumulated, the latent "thought" state continues to be updated. The final representation passed to $D$ is the output of ``thinking'' block $T$ after final iteration. \footnote{We also tried to follow \citet{Graves2016AdaptiveCT} to represent final representation as the weighted mixture of the outputs after each iteration, but found it less effective.}

The router is implemented as a linear projection of the hidden state followed by a sigmoid activation, initialized randomly. Despite its simplicity, this design proved effective in practice. The overall model is trained end-to-end with the standard task loss. Compared to a fixed-depth design, ACT introduces per-token dynamic depth, enabling more efficient and adaptive use of the recursive block.

\begin{figure}[h]
\begin{center}
%\framebox[4.0in]{$\;$}
% \fbox{\rule[-.5cm]{0cm}{4cm} \rule[-.5cm]{4cm}{0cm}}
\includegraphics[width=\textwidth]{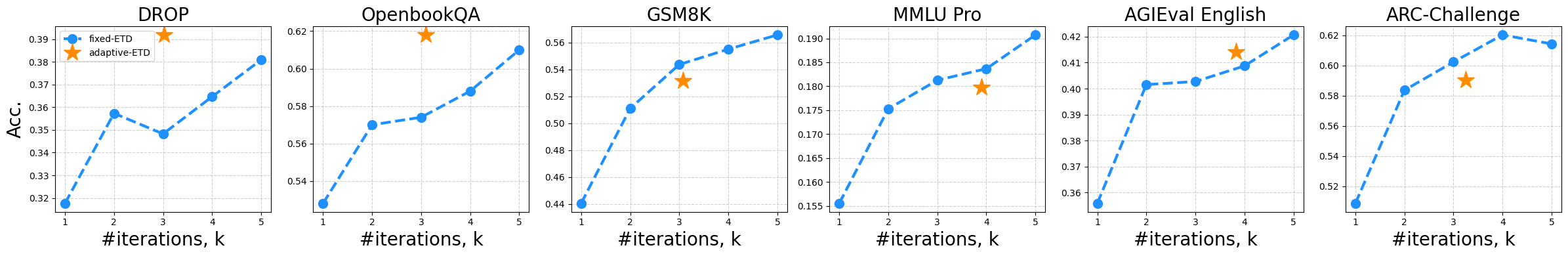}
\end{center}
\caption{Results of fixed-depth ETD with varying numbers of recursive ``thinking'' iterations compared to adaptive-depth ETD. For fixed-depth ETD, we report accuracy (Acc.) at each iteration count. For adaptive-depth ETD, we report accuracy and the average number of iterations per task.}
\label{fig:adaptive_plot}

\end{figure}

\subsection{Results}
We outlined the difference in architecture between fixed- and adaptive-depth approaches, while we follow the same training pipeline discussed in Section \ref{sec:training_pipeline}. Figure \ref{fig:adaptive_plot} reports the performance of fixed-depth ETD and adaptive-depth ETD,  together with the average number of loops per task.\footnote{We selected these tasks because they exhibit the largest relative gains from the recursive approach. See Appendix~\ref{app:etd_all_tasks} for results on the six tasks with the highest relative improvement of ETD ($k$=5) over baseline.} 

From Figure~\ref{fig:adaptive_plot}, we make three key observations. First, this exploratory approach in the direction of adaptive test-time compute approach shows clear improvement over baseline with no recursive iterations. Second, looking at the performance on DROP and OpenbookQA, both of which are reading comprehension tasks, we see that adaptive-depth ETD outperforms the ETD with fixed $k$=5 iterations. Moreover, it also achieves this with fewer iterations on average. Third, for the remaining tasks, adaptive-depth ETD follows the empirical accuracy–iteration tradeoff of the fixed-depth baselines. In particular, its accuracy matches the trend observed for increasing iteration counts, suggesting that performance is well-aligned with its average effective depth. Notably, in these tasks, the adaptive method halts additional iterations once further computation yields only marginal gains.

\section{Related Work}

\paragraph{\bf Recursive architectures}

Recurrence has long been a foundational concept, from RNNs to efforts to incorporate it into transformers. In transformers, recurrence has been explored by iteratively refining representations across all tokens in parallel \citep{Dehghani2018UniversalT, Lan2019ALBERTAL}, and applied to algorithmic tasks such as arithmetic \citep{Schwarzschild2021CanYL, Bansal2022EndtoendAS, Bear2024RethinkingDT, McLeish2024TransformersCD}. Other works offered theoretical and small-scale analyses of looped transformers \citep{Giannou2023LoopedTA, Gatmiry2024CanLT, Yang2023LoopedTA, Fan2024LoopedTF}.

Beyond fully recurrent-depth architectures, several hybrid designs have also been proposed, including latent sub-networks \citep{Li2020DeepTW}, Mixture-of-Experts structures \citep{Tan2023SparseUT, Csordas2024MoEUTMU}, and dynamic weight-tying \citep{Hay2024DynamicLT, Liu2024MobileLLMOS}. The major motivation of many works mentioned above was inspired by efficiency based on utilizing shared parameters.

\paragraph{\bf Latent Reasoning}
Chain-of-thought prompting has been a central focus in recent studies of reasoning \citep{Merrill2024TheEP, Feng2023TowardsRT, Li2024ChainOT}. In contrast, our proposal follows the alternative line of latent reasoning, where reasoning unfolds in the model’s hidden representations rather than explicit textual traces. Related efforts on learning to reason in continuous spaces include \citet{Hao2024TrainingLL, Cheng2024CompressedCO, Liu2024DeliberationIL, geiping2025scaling, saunshi2025reasoning}. \citet{Chen2024WhatCT, Ye2024PhysicsOL, petty2023impact} have shown the importance of model depth for reasoning. We step further showing that larger depth leads to reasoning improvements also when it is achieved via looping, without increasing the number of parameters.

\paragraph{\bf Adaptive Computation}
Dynamic compute allocation has been shown to substantially reduce training and inference costs, spanning from early neural networks \citep{Bengio2015ConditionalCI, Huang2016DeepNW, Teerapittayanon2016BranchyNetFI, Panda2015ConditionalDL}  to LLMs \citep{Hou2020DynaBERTDB, Elbayad2019DepthAdaptiveT, Fedus2021SwitchTS, Bae2023FastAR, Elhoushi2024LayerSkipEE}. A prominent line of work, early exiting, learns to terminate computation on “easy” inputs by skipping subsequent layers \citep{Elbayad2019DepthAdaptiveT, Schuster2022ConfidentAL, Bae2023FastAR, Elhoushi2024LayerSkipEE}.  Adaptive depth can be also formulated  as a routing problem: each layer’s router selects a subset of tokens for full computation while others bypass the layer, enabling token-level conditional compute \citep{Raposo2024MixtureofDepthsDA, Luo2024MoDEM}. Extending this idea, \cite{Bae2025MixtureofRecursionsLD} applied conditional routing to recursive transformers, but restricted recursion to a small, fixed maximum of three iterations. 

\paragraph{\bf Key Differences from Prior Work}
Our approach differs from prior work in several important ways. First, most recursive-depth methods have been studied primarily as a means of improving parameter efficiency \citep{Lan2019ALBERTAL, Bae2024RelaxedRT}, i.e., reducing parameter count while maintaining performance, whereas our focus is on enhancing reasoning capability. Second, to our knowledge, we are the first to propose a recursive approach guided by interpretability: rather than choosing the recursive configuration heuristically, we iterate specifically over layers critical for reasoning. Third, our method is simple and requires no additional components such as extra latent states for recursive blocks and very large of number of iterations \citep{geiping2025scaling}, LoRA adapters \citep{Bae2024RelaxedRT}, regularization terms \citep{saunshi2025reasoning}, or input injections \citep{Aleksandrov2025AbbIEAB}. Fourth, unlike most prior work that evaluated recurrence under simplified setups, we show that recursive depth improves advanced open-source models trained with state-of-the-art practices in architecture, training recipes, and pretraining mixtures, validating our approach extensively on real-world reasoning tasks. Speaking of adaptive-depth recursive model, in our formulation we advocate for open-ended test-time compute scaling: after each iteration, the model should autonomously decide whether to continue or halt, without being constrained by a predefined cap \citep{Bae2025MixtureofRecursionsLD}.

\section{Conclusions}

We introduced \textit{Encode–Think–Decode} (ETD), a paradigm that enhances the reasoning abilities of LLMs by performing latent-space reasoning. Unlike approaches that depend on scaling model size or externalizing reasoning through CoT prompting, ETD amplifies reasoning-relevant computations within the model itself, without altering its architecture, parameters, data, or hyperparameters. Across 17 benchmarks, ETD consistently improved performance, with substantial gains on reasoning-intensive tasks such as GSM8K and MATH. Our analysis underscores the importance of iterating over deeper, reasoning-relevant layers, and adaptive depth strategies further show how ETD can dynamically allocate compute based on task difficulty.

Overall, recursive latent reasoning emerges as a simple, effective, and broadly applicable approach for strengthening reasoning in LLMs. By integrating interpretability insights with recursive computation, ETD illustrates how leveraging depth and structure can advance reasoning in language models.

\section{Future Work}
Future work spans several directions. Extending ETD to multimodal  models could establish recursive latent reasoning as a general principle of representation learning across domains. Designing more efficient training strategies, together with refining adaptive depth mechanisms, may yield better compute–performance trade-offs. Assessing the impact of ETD on instruct models will require integration at the post-training stage, which we leave for future investigation. Last but not least, conducting interpretability studies could clarify how recursive latent reasoning interacts with model circuits and representations, offering deeper insights into the structure of reasoning in LLMs.

\section{Acknowledgments}
The authors thank Karen Hambardzumyan, Edan Toledo, Zheng Zhao, Preslav Aleksandrov, and Andrey Gromov for their insightful discussions and constructive feedback.

\clearpage
\newpage
\bibliographystyle{assets/plainnat}
\bibliography{paper}

\clearpage
\newpage
\beginappendix

\section{Computing Angular Distance}
\label{app:angular_distance}
Elaborating on the computation of angular distance in Section~\ref{sec:choosing_config}, 
the angular distance for a single sequence of length $T$ is defined as
\[
d\!\left(x^{(\ell)}, x^{(\ell+n)}\right) 
= \frac{1}{\pi} \arccos \left( 
\frac{x^{(\ell)}_{T} \cdot x^{(\ell+n)}_{T}}
{\|x^{(\ell)}_{T}\| \, \|x^{(\ell+n)}_{T}\|} 
\right),
\]
where the inner product is taken over the hidden dimension of the model for the last token $T$ of the sequence, $\|\cdot\|$ denotes the $L^{2}$ norm, and the factor $1/\pi$ normalizes the distance to $[0,1]$. 
We average this distance over 10,000 examples to obtain a stable estimate. 
We focus on the final token since, under a causal attention mask, its embedding is the only one that depends on the entire sequence. 
We use the same definition of angular distance as \cite{gromov2024unreasonable}.

\section{Detailed Evaluation Benchmarks}
\label{app:eval_benchmarks}

To capture broad conceptual nature of reasoning, we consider 17 real-world benchmarks grouped into six categories, ordered along a spectrum from less to more reasoning intensive tasks, i.e. from factual recall to systematic symbolic reasoning: factual knowledge, reading comprehension, commonsense reasoning,  multi-disciplinary Reasoning, BIG-Bench Hard (BBH), and mathematical reasoning. This progression reflects increasing reliance on reasoning rather than memorization.

\begin{itemize}

    \item \textbf{Factual Knowledge:} Tasks that test the model's ability to recall information without additional context, thus primarily measuring memorization. We include TriviaQA \citep{Joshi2017TriviaQAAL} and NaturalQuestions \citep{Kwiatkowski2019NaturalQA}.
    \item \textbf{Reading Comprehension:} Tasks requiring the model to infer answers from a given passage, involving text understanding and light reasoning (e.g., multi-hop). Benchmarks include BoolQ \citep{Clark2019BoolQET}, OpenBookQA \citep{Mihaylov2018CanAS}, and DROP \citep{Dua2019DROPAR}.
    \item \textbf{Commonsense Reasoning:} Tasks that evaluate human-like capacity to make assumptions and inferences about the nature and characteristics of everyday scenarios, including CommonSenseQA \citep{Talmor2019CommonsenseQAAQ}, HellaSwag \citep{Zellers2019HellaSwagCA}, SocialQA \citep{sap2019socialiqa}, WinoGrande \citep{sakaguchi2021winogrande}. 
    \item \textbf{Multi-Disciplinary Reasoning:} Benchmarks testing both factual knowledge and reasoning across broad academic and multi-disciplinary domains. We include ARC-Easy and ARC-Challenge \citep{Clark2018ThinkYH}, MMLU \citep{Hendrycks2020MeasuringMM}, MMLU-Pro \citep{Wang2024MMLUProAM}, and AGIEval-English \citep{Zhong2023AGIEvalAH}.

    \item \textbf{BIG-Bench Hard (BBH):} A collection of 23 diverse tasks spanning math, logic puzzles, symbolic and social reasoning \citep{Suzgun2022ChallengingBT}. Many tasks are synthetic, and BBH serves as a cross-cutting benchmark for compositional reasoning that does not fit neatly into the other categories.
    \item \textbf{Mathematical Reasoning:} We finally test the model on solve math word problem benchmarks to evaluate systematic reasoning and symbolic manipulation, represented by GSM8K \citep{Cobbe2021TrainingVT} and MATH \citep{Hendrycks2021MeasuringMP}.
\end{itemize}

\section{Algorithm for choosing the optimal configuration}
\label{sec: kneedle_algorithm}

To automatically identify the point at which a curve transitions from a rapid to a gradual decrease, we employ the \textit{Kneedle algorithm} \citep{Satopaa2011FindingA}. The difference function $D_i$ is then evaluated on $(x, \tilde{y}(x))$, providing a smooth approximation that avoids spurious local variations.

Formally, let the curve be represented as a sequence of points:
\[
\mathcal{C} = \{ (x_i, y_i) \}_{i=0}^n,
\]
where $x$ corresponds to the layer index $l$ and $y$ to the angular distance $d(l, l+1)$.The key steps underlying Kneedle Algorithm are:
\begin{enumerate}

    \item Smooth and normalize the data into $[0,1]^2$: ($\hat{x}_i$, $\hat{y}_i$).
    \item Compute the deviation $D_i = \hat{y}_i - (1-\hat{x}_i)$ from the diagonal.
    \item Identify local maxima of the difference curve as candidate knees.
    \item Apply a threshold-based rule (with sensitivity parameter $S$) to
          declare knees when the difference drops below threshold.
\end{enumerate}

% \noindent
To improve robustness against noise, we apply a polynomial interpolation of degree~2 to the data:
\[
\tilde{y}(x) = a_0 + a_1 x + a_2 x^2,
\]
fitted via least squares. This provides a smooth approximation that avoids spurious local variations.

The details of Kneedle Algorithm can be summarized as follows:

\begin{enumerate}
    \item Normalization: Scale both axes to $[0,1]$:
    \[
    \hat{x}_i = \frac{x_i - \min(x)}{\max(x) - \min(x)}, 
    \qquad
    \hat{y}_i = \frac{y_i - \min(y)}{\max(y) - \min(y)}.
    \]

    \item Difference curve: Compute the deviation between the normalized curve and the diagonal $y = 1 - \hat{x}$:
    \[
    D_i = \hat{y}_i - (1 - \hat{x}_i).
    \]

    \item Local maxima: Candidate knees are local maxima of $D_i$, i.e.
    \[
    D_{i-1} < D_i \quad \wedge \quad D_{i+1} < D_i.
    \]

    \item Threshold rule: For each local maximum, define a threshold
    \[
    T_i = D_i - S \cdot \Delta_x, 
    \quad \Delta_x = \tfrac{1}{n-1}\sum_{j=1}^{n-1} (\hat{x}_{j+1} - \hat{x}_j),
    \]
    where $S > 0$ is a sensitivity parameter. A knee is declared at $i^\ast$ if $D_j < T_i$ for some $j > i$ before the next local maximum is reached.
\end{enumerate}

\noindent
We run the above procedure using the \texttt{KneeLocator} package:
\begin{verbatim}
kneedle = KneeLocator(
    x, y,
    curve='convex',
    direction='decreasing',
    interp_method='polynomial',
    polynomial_degree=2,
    online=True
)
\end{verbatim}
The returned index
\[
i^\ast = \texttt{kneedle.knee}
\]
is taken as the transition point from steep to gradual decline.

\section{Performance of ETD on each task}
\label{app:etd_all_tasks}

Table \ref{tab:looping_results} reports the results of the Encode–Think–Decode (ETD) method with varying numbers of iterations over recursive “thinking” blocks, compared to the OLMo 2 1B baseline on 6 categories of tasks described in Sec. \ref{sec:evaluation_metrics}. In this section, we share the results for each individual tasks in Tables \ref{tab:looping_results_tasks_1}.

\begin{table}[ht!]
\caption{Results of the Encode–Think–Decode (ETD) method with varying numbers of iterations over recursive “thinking” blocks, compared to the OLMo 2 1B baseline. Reported metrics include accuracy (Acc.) and relative improvement ($\Delta$, in \%) with respect to the baseline. Parameter counts denote the number of distinct layers, while FLOPs correspond to the number of effective forward-pass layers.}

\label{tab:looping_results_tasks_1}
\centering
\renewcommand{\arraystretch}{1.2}
\setlength{\tabcolsep}{5pt} 

\resizebox{\textwidth}{!}{%
\begin{tabular}{c c| *{5}{c c|}c c}

\toprule
\noalign{\vskip 0.5ex}  
& 
& 
\multicolumn{2}{c}{\makecell{Natural Questions}} & 
\multicolumn{2}{c}{\makecell{TriviaQA}} & 
\multicolumn{2}{c}{\makecell{BoolQ}} & 
\multicolumn{2}{c}{\makecell{OpenbookQA}} & 
\multicolumn{2}{c}{DROP} & 
\multicolumn{2}{c}{\makecell{HellaSwag}} \\
% [0.75ex] 
\cmidrule(lr){3-4} \cmidrule(lr){5-6} \cmidrule(lr){7-8} \cmidrule(lr){9-10} \cmidrule(lr){11-12} \cmidrule(lr){13-14}
Model & Params/FLOPs & \multicolumn{1}{c}{Acc.} & $\Delta$ & Acc. & $\Delta$ & Acc. & $\Delta$ & Acc. & $\Delta$ & Acc. & $\Delta$ & Acc. & $\Delta$ \\
\midrule
Baseline & 16 / 16  & 20.98 & - & 54.12 & - & 72.0 & - & 52.8 & - & 31.761 & - & 69.7 & - \\
\rowcolor{gray!10}
Ours (k=2) & 16 / 20 & 20.76 & (-1.01\%) & 55.43 & (+2.43\%) & 75.7 & +(5.14\%) & 57.0 & (+7.95\%) & 35.73 & (+12.5\%) & 69.8 & (+0.14\%) \\
\rowcolor{gray!5}
Ours (k=3) & 16 / 24 & 19.97 & (-4.78\%) & 55.13 & (+1.88\%) & 76.0 & (+5.56\%) & 57.4 & (+8.71\%) & 34.82 & (+9.64\%) & 69.6 & (-0.14\%) \\
\rowcolor{gray!10}
Ours (k=4) & 16 / 28  & 20.35 & (-2.99\%) & 55.13 & (+1.8\%)8 & 78.0 & (+8.33\%) & 58.8 & (+11.36\%) & 36.47 & (+14.81\%) & 71.0 & (+1.87\%) \\
\rowcolor{gray!5}
Ours (k=5) & 16 / 32 & 20.53 & (-2.12\%) & 55.93 & (+3.36\%) & 76.4 & (+6.11\%) & 61.0 & (+15.53\%) & 38.086 & (+19.91\%) & 70.4 & (+1\%) \\
\bottomrule
\end{tabular}%
}
% \end{table}

% \begin{table}[ht!]
% \caption{Results}
\vspace{10pt}
% \label{tab:looping_results_tasks_2}
\centering
\renewcommand{\arraystretch}{1.2}
\setlength{\tabcolsep}{5pt} % reduce padding

\resizebox{\textwidth}{!}{%
% \begin{tabular}{c c|*{5}{c c|}c c}
\begin{tabular}{c c| *{5}{c c|}c c}

\toprule
\noalign{\vskip 0.5ex}  % adds vertical space after the line
% \multicolumn{1}{c}{\multirow{2}{*}{Model}} 
& 
% \multicolumn{1}{c}{\multirow{1}{*}{\makecell{Params/FLOPs}}} 
& 
% \multirow{2}{*}{FLOPs} & 
\multicolumn{2}{c}{\makecell{Social IQa}} & 
\multicolumn{2}{c}{\makecell{WinoGrande}} & 
\multicolumn{2}{c}{\makecell{CommonsenseQA}} & 
\multicolumn{2}{c}{\makecell{ARC-Easy}} & 
\multicolumn{2}{c}{ARC-Challenge} & 
\multicolumn{2}{c}{\makecell{MMLU}} \\
% [0.75ex] 
\cmidrule(lr){3-4} \cmidrule(lr){5-6} \cmidrule(lr){7-8} \cmidrule(lr){9-10} \cmidrule(lr){11-12} \cmidrule(lr){13-14}
Model & Params/FLOPs & \multicolumn{1}{c}{Acc.} & $\Delta$ & Acc. & $\Delta$ & Acc. & $\Delta$ & Acc. & $\Delta$ & Acc. & $\Delta$ & Acc. & $\Delta$ \\
\midrule
Baseline & 16 / 16  & 58.1 & - & 66.69 & - & 66.67 & - & 78.5 & - & 50.85 & - & 44.52 & - \\
\rowcolor{gray!10}
Ours (k=2) & 16 / 20 & 62.9 & (+8.26\%) & 66.85 & (+0.24\%) & 67.40 & (+1.11\%) & 78.4 & (-0.13\%) & 58.36 & (+14.77\%) & 47.59 & (+6.9\%) \\
\rowcolor{gray!5}
Ours (k=3) & 16 / 24 & 63.9 & (+9.98\%) & 68.19 & (+2.25\%) & 69.29 & (+3.93\%) & 79.7 & (+1.53\%) & 60.24 & (+18.46\%) & 49.40 & (+10.96\%) \\
\rowcolor{gray!10}
Ours (k=4) & 16 / 28  & 65.0 & (+11.88\%) & 68.51 & (+2.72\%) & 68.14 & (+2.21\%) & 79.8 & (+1.66\%) & 62.03 & (+21.98\%) & 49.84 & (+11.95\%) \\
\rowcolor{gray!5}
Ours (k=5) & 16 / 32 & 66.2 & (+13.94\%) & 68.59 & (+2.84\%) & (+68.47\%) & (+2.7\%) & 80.4 & (+2.42\%) & 61.43 & (+20.81\%) & 49.95 & (+12.19\%) \\
\bottomrule
\end{tabular}%
}
% \end{table}

% \begin{table}[ht!]
% \caption{Results}
\vspace{10pt}

% \label{tab:looping_results_tasks_3}
\centering
\renewcommand{\arraystretch}{1.2}
\setlength{\tabcolsep}{5pt} % reduce padding

\resizebox{\textwidth}{!}{%
% \begin{tabular}{c c|*{5}{c c|}c c}
\begin{tabular}{c c| *{4}{c c|}c c}

\toprule
\noalign{\vskip 0.5ex}  % adds vertical space after the line
% \multicolumn{1}{c}{\multirow{2}{*}{Model}} 
& 
% \multicolumn{1}{c}{\multirow{1}{*}{\makecell{Params/FLOPs}}} 
& 
% \multirow{2}{*}{FLOPs} & 
\multicolumn{2}{c}{\makecell{MMLU Pro}} & 
\multicolumn{2}{c}{\makecell{AGIEval English}} & 
\multicolumn{2}{c}{\makecell{BBH}} & 
\multicolumn{2}{c}{\makecell{GSM8K}} & 
\multicolumn{2}{c}{MATH}  \\
% [0.75ex] 
\cmidrule(lr){3-4} \cmidrule(lr){5-6} \cmidrule(lr){7-8} \cmidrule(lr){9-10} \cmidrule(lr){11-12} %\cmidrule(lr){13-14}
Model & Params/FLOPs & \multicolumn{1}{c}{Acc.} & $\Delta$ & Acc. & $\Delta$ & Acc. & $\Delta$ & Acc. & $\Delta$ & Acc. & $\Delta$ \\
\midrule
Baseline & 16 / 16  & 15.55 & - & 35.58 & - & 31.8 & - & 44.05 & - & 4.57 & - \\
\rowcolor{gray!10}
Ours (k=2) & 16 / 20 & 17.53 & (12.72\%) & 40.16 & (12.86\%) & 31.67 & (-0.4\%) & 51.10 & (+16.01\%) & 5.45 & (19.22\%) \\
\rowcolor{gray!5}
Ours (k=3) & 16 / 24 & 18.13 & (+16.57\%) & 40.27 & (+13.2\%) & 32.62 & (+2.58\%) & 54.36 & (+23.41\%) & 6.22 & (+36.04\%) \\
\rowcolor{gray!10}
Ours (k=4) & 16 / 28  & 18.37 & (+18.12\%) & 40.88 & (+14.89\%) & 33.01 & (+3.82\%) & 55.50 & (+25.99\%) & 3.73 & (-18.28\%) \\
\rowcolor{gray!5}
Ours (k=5) & 16 / 32 & 19.07 & (+22.66\%) & 42.07 & (+18.24\%) & 33.49 & (+5.3\%) & 56.56 & (+28.4\%) & 4.33 & (-5.17\%) \\
\bottomrule
\end{tabular}%
}
\end{table}

\section{Results with iterations over different layers}
\label{app:looping_layers}

We fix the recursive ``thinking'' block size and vary its starting position from layer 2 to 12 in steps of 2, which is equivalent to increasing the size of the latent encoder $N_{E}$ from 1 to 11 in steps of 2.

\begin{table}[ht!]

\caption{Results of the Encode–Think–Decode (ETD) method when varying the subset of layers in the recursive block. We report accuracy (Acc.) when increasing the size of the latent encoder $N_{E}$ from 1 to 11 in steps of 2, for each of six task categories (as defined in Sec. \ref{sec:evaluation_metrics}).}

\label{tab:looping_layers}
\centering
\renewcommand{\arraystretch}{1.2}
\setlength{\tabcolsep}{4pt} 

\resizebox{\textwidth}{!}{
\begin{tabular}{c|c|*{5}{c |}c}
\toprule
\noalign{\vskip 0.5ex}  
\multirow{1}{*}{Model} & 
\multirow{1}{*}{\makecell{Params/\\FLOPs}} & 
% \multirow{2}{*}{FLOPs} & 
\multicolumn{1}{c|}{\makecell{Factual\\Knowledge}} & 
\multicolumn{1}{c|}{\makecell{Reading\\Comprehension}} & 
\multicolumn{1}{c|}{\makecell{Commonsense\\Reasoning}} & 
\multicolumn{1}{c|}{\makecell{Multi-Disciplinary\\ Reasoning}} & 
\multicolumn{1}{c|}{BBH} & 
\multicolumn{1}{c}{\makecell{Math.\\Reasoning}} \\
[1.75ex] 
\midrule

\rowcolor{gray!10}
1-4*2-11 & 16 / 20  & 37.92  & 55.53  & 64.82  & 44.99  & 31.23 & 25.6  \\
\rowcolor{gray!5}
3-4*2-9 & 16 / 20 & 37.43  & 56.93  & 65.87  & 46.9  & 29.80  & 27.31  \\
\rowcolor{gray!10}
5-4*2-7 & 16 / 20 & 37.58  & 56.51  & 66.86 & 49.03  & 32.21  & 26.8  \\
\rowcolor{gray!5}
7-4*2-5 & 16 / 20  & 38.1  & 56.14  & 66.74  & 48.41  & 31.67  & 28.27  \\
\rowcolor{gray!10}
9-4*2-3 & 16 / 20 & 37.7  & 53.46  & 65.52  & 45.71  & 31.05 & 27.35 \\
\rowcolor{gray!5}
11-4*2-1 & 16 / 20 & 37.67  & 54.79  & 64.45  & 45.18  & 30.93 & 24.63 \\
\bottomrule
\end{tabular}%
}
\end{table}

\end{document}